\documentclass[10pt,twocolumn,letterpaper]{article}

\usepackage{cvpr}              %

\definecolor{cvprblue}{rgb}{0.21,0.49,0.74}
\usepackage[pagebackref,breaklinks,colorlinks,allcolors=cvprblue]{hyperref}
\usepackage[capitalize,noabbrev,nameinlink]{cleveref}

\usepackage{makecell} 
\usepackage[utf8]{inputenc} %
\usepackage[T1]{fontenc}    %
\usepackage{url}            %
\usepackage{amsfonts}       %
\usepackage{nicefrac}       %
\usepackage{microtype}      %
\usepackage{xcolor}         %
\usepackage{color}
\usepackage{longtable}
\usepackage{threeparttable}
\usepackage{xspace}
\usepackage{appendix}
\usepackage{booktabs}
\usepackage{textcomp}
\usepackage{amsmath, amssymb, amsthm}
\usepackage{algorithm}
\usepackage[noend]{algpseudocode}
\algtext*{EndProcedure}%
\usepackage{xcolor,colortbl}
\usepackage{footnote}
\usepackage{tabularx}
\usepackage{multicol}
\usepackage{multirow}
\usepackage{pifont}
\usepackage{bm}
\usepackage{relsize}
\usepackage{mathtools}
\usepackage{bbm}
\usepackage{blindtext}
\usepackage{balance}
\usepackage{wrapfig}
\usepackage{caption}
\usepackage{subcaption}
\newcommand{\mypar}[1]{\vspace{0.5mm}\noindent\textbf{#1}}
\usepackage{pifont}
\usepackage{enumerate}

\usepackage{wasysym}
\usepackage{amssymb}%
\usepackage{pifont}%
\usepackage{mdframed}
\usepackage{float}

\newcommand{\method}{\textsc{MM-Graph}\xspace}

\newcommand{\cmark}{\ding{51}}%
\def\etal{\emph{et al.}}
\newcommand{\failCell}{\xmark}
\renewcommand{\failCell}{}
\newcommand{\winCell}{{\color{green}{\cmark}}}

\definecolor{Gray}{gray}{0.85}
\definecolor{msred}{rgb}{0.753,0.314,0.302}

\definecolor{lblue}{rgb}{0.3, 0.6, 1.0}
\definecolor{lgreen}{rgb}{0.908, 0.961, 0.908}
\definecolor{lblue}{rgb}{0.9, 0.92, 1.0}
\definecolor{lred}{rgb}{1.0, 0.885, 0.885}
\definecolor{lorange}{rgb}{0.960, 0.866, 0.721}

\newcommand{\numstd}[2]{#1 {\scriptsize$\pm$#2}}

\def\paperID{17144} %
\def\confName{CVPR}
\def\confYear{2025}

\title{Mosaic of Modalities: A Comprehensive Benchmark for Multimodal Graph Learning}

\author{
Jing Zhu$^{1}$,  Yuhang Zhou$^{3}$, Shengyi Qian$^{1}$, Zhongmou He$^{1}$, Tong Zhao$^{2}$, 
Neil Shah$^{2}$, Danai Koutra$^{1}$\\
  $^1$University of Michigan\quad $^2$Snap Inc.\quad $^3$University of Maryland  \\
  \texttt{\small\url{https://mm-graph-benchmark.github.io/}}\\
}

\makeatletter
\g@addto@macro\@maketitle{
\vspace{-3.7em}
\begin{figure}[H]
   \setlength{\linewidth}{\textwidth}
\setlength{\hsize}{\textwidth}
\centering
\includegraphics[width=\linewidth]{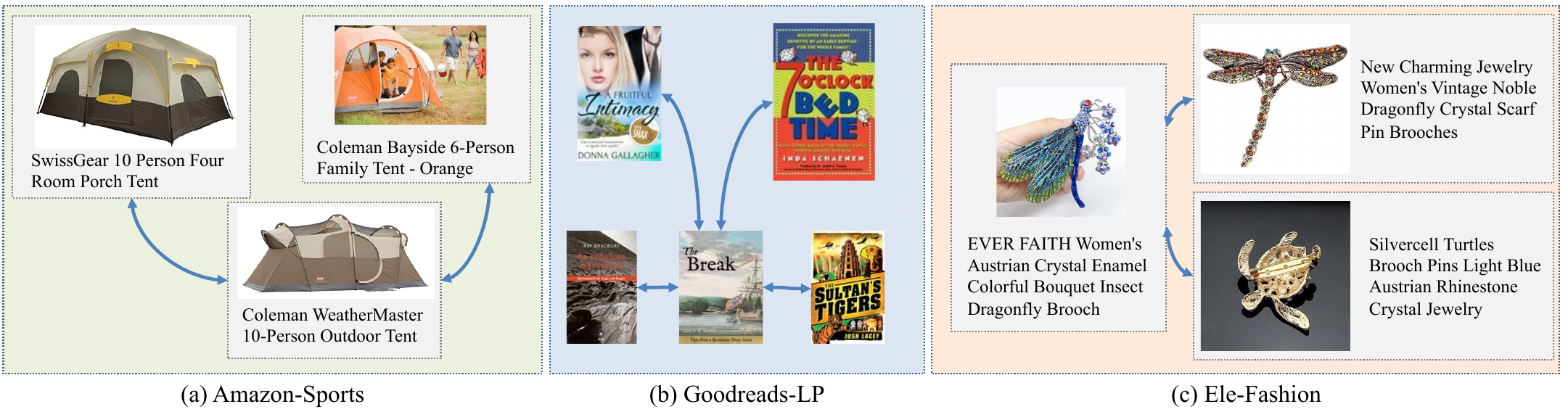}
\vspace{-2em}
\caption{Visualization of our Multimodal Graph Benchmark (\method). All nodes of our benchmark have both visual and text features. \textbf{(a) Amazon-Sports:} The image and text come from the original image and title of the sports equipment. \textbf{(b) Goodreads-LP:} The images correspond to book covers. We do not show the text features of Goodreads-LP since the book description is very long. \textbf{(c) Ele-fashion:} The images and texts correspond to the original image and title of the fashion product, respectively.
}
\label{fig:teaser}
\end{figure}
}
\makeatother

\begin{document}

\maketitle

\begin{abstract}

Graph machine learning has made significant strides in recent years, yet the integration of visual information with graph structure and its potential for improving performance in downstream tasks remains an underexplored area. To address this critical gap, we introduce the Multimodal Graph Benchmark (\method), a pioneering benchmark that incorporates both visual and textual information into graph learning tasks. \method extends beyond existing text-attributed graph benchmarks, offering a more comprehensive evaluation framework for multimodal graph learning.
Our benchmark comprises seven diverse datasets of varying scales (ranging from thousands to millions of edges), designed to assess algorithms across different tasks in real-world scenarios. These datasets feature rich multimodal node attributes, including visual data, which enables a more holistic evaluation of various graph learning frameworks in complex, multimodal environments.
To support advancements in this emerging field, we provide an extensive empirical study on various graph learning frameworks when presented with features from multiple modalities, particularly emphasizing the impact of visual information. This study offers valuable insights into the challenges and opportunities of integrating visual data into graph learning.

\end{abstract}

\section{Introduction}
\label{sec:intro}

Graphs are powerful data structures that capture complex relationships across various domains, from social networks to recommendation systems. In real-world applications, these graph entities often possess rich, multimodal semantic information, including crucial visual data. The integration of visual information with structured graph data is particularly vital for tasks such as relevance search and information retrieval in recommendation systems~\cite{su2021detecting, crandall2009mapping, guo2011role}.

Existing graph learning benchmarks have made significant strides in evaluating graph neural networks (GNNs) on various connectivity patterns and topologies~\cite{dwivedi2022long, gui2022good, li2024gslb, morris2020tudataset, hu2020open}, they often overlook the critical role of visual information. Moreover, recent advancements in language modeling have led to effective integration of textual information with graph structures, resulting in comprehensive benchmarks for text-attributed graphs~\cite{yan2023comprehensive, zhu2023touchup, jin2023large, peng2024learning}. However, visual information offers unique and invaluable insights that text and structure alone cannot capture~\cite{wei2024rendering,wei2024gita}. Visual data can reveal similarities and relationships in artist styles that may not be apparent from textual or structural information alone. For instance, as illustrated in \cref{fig:teaser}, visually similar products may have distinct textual descriptions, creating a semantic gap that text-only GNNs might fail to bridge. By incorporating visual information, graph learning algorithms can better understand and exploit the intricate relationships between multimodal semantics and graph structures. This visual context can significantly enhance the performance of GNNs. Integrating visual data into graph learning frameworks enables a more holistic understanding of the underlying relationships in complex, real-world networks. It allows GNNs to capture nuanced information that may be missed when relying solely on textual or structural data, leading to more accurate and contextually relevant predictions.

To address the critical need for integrating visual data in graph machine learning, we introduce the Multimodal Graph Benchmark (\method), the first comprehensive graph benchmark that explicitly incorporates visual information alongside textual data. \method comprises seven diverse graph learning datasets of varying scales, specifically designed to evaluate two fundamental categories of graph tasks: node and link property predictions. These tasks require models to make predictions at different levels of graph granularity, leveraging the rich multimodal information available. A key feature of \method is its inclusion of high-resolution visual node features, which sets it apart from existing benchmarks. Each dataset within \method contains not only textual information and traditional graph structures but also associated images for nodes. This visual data provides crucial additional context that can significantly enhance the performance of graph learning algorithms in real-world scenarios.

While multimodal knowledge graphs (MMKGs) incorporating visual information exist, the community faces challenges with limited datasets, often with outdated or inaccessible image URLs and poor dataset quality~\cite{liu2019mmkg, safavi2020codex, liang2024survey, chen2024knowledge}. Recognizing the critical need for high-quality MMKG datasets that effectively integrate visual data~\cite{liang2024survey}, our MMKGs are sourced from CoDEx, a recent comprehensive knowledge graph benchmark known for its diverse and interpretable content and challenging link prediction tasks~\cite{safavi2020codex}. By enhancing CoDEx with curated visual information, our MMKGs offer higher quality datasets that address a significant gap in the MMKG community, providing a robust foundation for evaluating how graph learning models incorporate and benefit from visual features in knowledge graph scenarios. We highlight the key differences between \method and existing benchmarks in ~\cref{tab:salesman}.

\begin{table}[t]
\caption{Qualitative comparison of existing benchmarks. Abbreviations: \underline{LP} = Link Prediction. \underline{NC} = Node Classification. \underline{KGC} = Knowledge Graph Completion.}
\label{tab:salesman}
\centering

\resizebox{0.9\columnwidth}{!}{
\begin{tabular}{l|cccccc}
\toprule
  Benchmarks & Language &  Vision &  NC & LP & KGC \\  
\midrule
GL-Bench~\cite{li2024glbench} & \winCell & \failCell & \winCell & \failCell & \failCell \\
CS-TAG~\cite{yan2023comprehensive} & \winCell & \failCell & \winCell & \winCell & \failCell \\
VTKG~\cite{lee2023vista} & \winCell & \winCell & \failCell & \failCell & \winCell \\
CoDEx~\cite{safavi-koutra-2020-codex} & \winCell & \failCell & \failCell & \failCell & \winCell \\
OGB~\cite{hu2020open} & \failCell & \failCell & \winCell & \winCell & \winCell \\
\midrule
\method & \winCell & \winCell & \winCell & \winCell & \winCell \\
\bottomrule
\end{tabular}
}
\vspace{-0.3cm}
\end{table}

\noindent \textbf{Contributions and novelty.}  In this work, we introduce a multimodal graph benchmark known
as \method. \method fills this gap by providing a diverse set of datasets with varying scales, tasks, and modality combinations. By establishing a common ground for evaluation, \method enables researchers to reflect on the strengths and limitations of existing multimodal graph learning algorithms and identify areas for improvement. As design principles, we strive to create non-trivial, realistic data that comes from real-world applications and can benefit real-world deployments. Specifically, \method contains three link prediction datasets, two node classification datasets, and two knowledge graph completion datasets as shown in \cref{tab:datasets-overview}. We summarize our novel contributions as follows:

(1) \textbf{Text- and Image- Attributed Graph(TIAG) Benchmarks.} First standardized benchmark providing both textual and visual features across various graph learning tasks, enabling a more holistic understanding of complex data relationships.

(2) \textbf{Benchmarking Multimodal GNNs.} Comprehensive performance analysis of diverse methods on \method, highlighting the need for more advanced multimodal Graph Neural Networks (GNNs).

(3) \textbf{Exploring Multimodal Feature Encodings.}
Pioneering investigation of various node feature encoding strategies (e.g. ImageBind), addressing the challenge of optimal feature representation in graph learning tasks.

\begin{table*}[t!]
\caption{Overview of the datasets in our proposed \method benchmark. We present five multimodal graph datasets with varying scale and tasks. All datasets are split at random. Abbreviations: \underline{LP} = Link Prediction. \underline{NC} = Node Classification. \underline{KGC} = Knowledge Graph Completion. \underline{OR} = Original.}
\label{tab:datasets-overview}
\centering
\setlength{\tabcolsep}{10pt} 
{\small
\begin{tabular}{lrrrrrr}
\toprule

\textbf{Name} & \textbf{Text Features} & \textbf{Visual Features} & \makecell{ \textbf{Task}} & \textbf{Metrics} & \textbf{Scale} & \textbf{Split Ratio}\\
\midrule

Amazon-Sports & Product Titles & Product Images & LP & MRR, Hits@K & Small & 8/1/1 \\

Amazon-Cloth & Product Titles & Product Images & LP & MRR, Hits@K & Medium & 8/1/1 \\

Goodreads-LP & Book Description & Book Images & LP & MRR, Hits@K & Large & 6/1/3\\

Ele-fashion & Fashion Titles& Fashion Images & NC & Accuracy & Medium & 6/1/3
\\

Goodreads-NC & Book Description & Book Images & NC & Accuracy & Large & 6/1/3
\\

MM-CoDEx-s & Entity Description & Entity Images & KGC & MRR, Hits@K & Small & OR
\\

MM-CoDEx-m & Entity Description & Entity Images & KGC & MRR, Hits@K & Medium & OR
\\

\bottomrule
\end{tabular}
}
\end{table*}

\section{Related Work}
\label{sec:related}
\noindent \textbf{Multimodal Feature Learning.} Multimodal feature learning aims to learn joint representations from multiple modalities, such as text, images, and audio. Transformer-based models have shown remarkable success in multimodal feature learning ~\cite{xu2023multimodal, radford2021learning}. These models learn typically transferable visual representations by leveraging corresponding natural language supervision. Models like FLAVA ~\cite{singh2022flava} and Perceiver~\cite{jaegle2021perceiver} have demonstrated the effectiveness of jointly pre-training transformers on unpaired images and text while CLIP has shown that contrastive objectives can effectively align representations from different modalities ~\cite{singh2022flava,jaegle2021perceiver,radford2021learning}. We refer to the survey for more details ~\cite{xu2023multimodal}.

 \noindent \textbf{Multimodal Graph Learning.} While most existing multimodal learning approaches focus on 1-to-1 mappings between modalities, real-world data often exhibits more complex many-to-many relationships that can be represented as graphs. Multimodal graph learning (MMGL) aims to address this challenge by leveraging graph structure to learn from multimodal data with complicated relations among multiple multimodal neighbors ~\cite{ektefaie2023multimodal, yoon2023multimodal, fang2025graphgpt, he2025unigraph2}. Recent efforts in MMGL have focused on integrating GNNs/Knowledge Graph Embeddings (KGEs) with pretrained language models (LMs)~\cite{yan2023comprehensive}. Our work builds upon the recent advances in multimodal feature learning and MMGL by introducing a standardized benchmark that incorporates both structured and unstructured modalities.

\noindent \textbf{Benchmarks for graph representation learning.} Several established graph benchmarks have been developed and widely adopted~\cite{morris2020tudataset, hu2020open, freitas2021large}. However, when it comes to learning on graphs with rich features, these
benchmarks exhibit notable deficiencies. Firstly, these datasets suffer from the absence of raw features, limiting the investigation of attribute modeling. Secondly, these datasets often neglect to explore the impact of feature modeling.  Thus, there is a compelling necessity to construct a comprehensive graph benchmark with rich natural features. Yan \etal{} proposed a graph benchmark with rich textual information, fostering the development of integrating GNNs with prevalent language models~\cite{yan2023comprehensive}. However, we argue that it is critical to investigate modalities other than text, especially images in structure learning tasks and proposing the first standarized multimodal graph benchmark to foster the development of research in this direction. The key differences between \method and existing benchmarks are highlighted in ~\cref{tab:salesman}.

\section{\method: Multimodal Graph Benchmark}
\label{sec:method}

In this section, we present the proposed 7 real-world datasets in \method. The general information about them is given in ~\cref{tab:datasets-overview}. We first introduce the data curation process of all datasets, as well as their license. For all datasets described below, we use Beautiful Soup to crawl images associated with the nodes in the graphs~\cite{richardson2007beautiful}.

\subsection{Data Curation}
~\label{sec:license}

\mypar{\textsc{Amazon-Sports/Amazon-Cloth}} are link prediction datasets, based on the Amazon-Review dataset~\cite{ni2019justifying, hou2024bridging}, where each node corresponds to a product on Amazon in the sports category and the link captures whether two products are co-purchased together. The text features are the titles of the products and the visual features are the high-resolution raw images of the products. In \cref{fig:teaser}(a), we show an example of nodes and edges in Amazon-Sports. The visual features are images of tents of various shapes under diverse backgrounds and text features are the titles of the tents. We preprocess the data by extracting the meta information of each item as well as their co-purchasing information. 

\mypar{\textsc{Goodreads-LP/Goodreads-NC}} are based on the Goodreads Book Graph dataset ~\cite{wan2018item, wan2019fine}. Here we construct the graph such that each node corresponds to a book on Goodreads and the link captures if a user who likes this book will like the other book, following ~\cite{zhu2023touchup}. The text features are the descriptions of the books and the visual features are the the book covers. Nodes without images are removed.

\mypar{\textsc{Ele-fashion}} is a node classification dataset, based on the Amazon-Fashion dataset~\cite{ni2019justifying, hou2024bridging}, where each node corresponds to a product and a link captures that the two products are copurchased. The text features are the titles of the products and the visual features are the images of the products. 

\mypar{\textsc{MM-CoDEx-s/MM-CoDEx-m}} is a knowledge graph completion dataset, based on the CoDEx-s dataset~\cite{safavi2020codex}. 
The text features come from the description of the Wikipedia article. For visual features, we crawl the corresponding images of the entity (e.g., person/place/item) from Wikipedia.

\mypar{Data Availability and Ethics.}
Our benchmark is organized from existing open source data, with proper open source licenses. 
\textsc{Amazon-Sports} and \textsc{Amazon-Cloth} and \textsc{Ele-Fashion} are available with Apache License~\footnote{\url{https://github.com/PeterGriffinJin/Patton}}.
\textsc{Goodreads-LP}, \textsc{Goodreads-NC}, \textsc{MM-CODEX-S} and \textsc{MM-CODEX-M} are released under MIT License~\footnote{\url{https://mengtingwan.github.io/data/goodreads.html}, \url{https://github.com/tsafavi/codex/tree/master}}.
These do not involve interaction with humans or private data.

\begin{table*}[t!]
\caption{Detailed graph-based statistics of datasets proposed in \method. 
\underline{CC} = Cluster Coefficient. \underline{RA} = Resource Allocation. \underline{N/A} = Not Applicable (Nodes do not have class labels). Detailed definition for each graph statistics or property are provided in the Appendix.}
\label{tab:datasets-stats}
\centering
\resizebox{0.88\textwidth}{!}
{
\begin{tabular}{l@{\hskip5pt}r@{\hskip5pt}r@{\hskip5pt}r@{\hskip5pt}r@{\hskip5pt}r@{\hskip5pt}r@{\hskip5pt}r}
\toprule
   \textbf{Name} &  \textbf{Nodes} & \textbf{Edges}  & \textbf{Average Degree} & \textbf{Average CC}  & \textbf{Average RA} & \textbf{Transitivity} & \textbf{Edge Homophily}\\
\midrule 

     Amazon-Sports & 50,250 &  356,202 & 14.18  &  0.4002 &  0.3377 & 0.2658 & N/A\\ 

    Amazon-Cloth &  125,839 & 951,271 &15.12 & 0.2940 &  0.2588  & 0.1846 & N/A\\

   Goodreads-LP & 636,502 & 3,437,017 & 10.79 & 0.1102 & 0.0685 & 0.0348 & N/A \\ 
   
    Goodreads-NC & 685,294 & 7,235,084 & 21.11 & 0.1614 &  0.1056 & 0.0498 & 0.6667
    \\

    Ele-fashion & 97,766 & 199,602 & 4.08 & 0.1730 & 0.1467& 0.0560 & 0.7675
    \\

\bottomrule
\end{tabular}

\vspace{-2em}
}
\end{table*}

\begin{table*}[t!]
\caption{Detailed MMKG statistics of datasets proposed in \method. (+): Positive (true) triples. (-): Verified negative (false) triples.}
\label{tab:kg-stats}
\centering
\resizebox{0.8\textwidth}{!}
{
\begin{tabular}{l@{\hskip 15pt}r@{\hskip15pt}r@{\hskip15pt}r@{\hskip15pt}r@{\hskip15pt}r@{\hskip15pt}r@{\hskip15pt}r}
\toprule
   \textbf{Name} &  \textbf{Entities} & \textbf{Relations}  & \textbf{Train (+)} & \textbf{Valid (+)}  & \textbf{Test (+)} & \textbf{Valid (-)} & \textbf{Test (-)}\\
\midrule 

    MM-CoDEx-s & 1,383 & 39 & 14,298 & 784 & 
 802 & 1028 & 1074
    \\

    MM-CoDEx-m & 7,697 & 51 & 47,617 & 2,628 & 2,595 & 4,721 & 4,746
    \\
\bottomrule
\end{tabular}

\vspace{-1.2cm}
}
\end{table*}

\subsection{Tasks}

\mypar{Link Prediction.}
The task is to predict new association edges given the training edges. The evaluation
is based on how well a model ranks positive test edges over negative test edges.  Specifically, for Amazon-Sports and Amazon-Cloth, we generate hard valid/test negatives using HeaRT~\cite{li2024evaluating}. HeaRT is recognized as a better way of generating negatives for link prediction in the community.  For Goodreads-LP, since it takes more than 120 hours to generate hard negatives using HeaRT, we perform random sampling. Each  positive edge in the validation/test set is ranked against 150 hard negative edges. 
For \textit{evaluation metrics}, we report MRR, Hits@10, and Hits@1, the three most commonly-used evaluation metrics for link prediction~\cite{hu2020open, li2024evaluating}.
For Amazon-Sports and Amazon-Cloth, 
edges are randomly split into train/valid/test splits according to 8/1/1 ratio. For Goodreads-LP, 
edges are randomly split according to 6/1/3 ratio. Validation and test edges are explicitly removed from the graphs to avoid any potential leakage~\cite{zhu2024pitfalls}.

\mypar{Node Classification.}
For Goodreads-NC, the task is to predict the category of items from 10 available categories, such as History, Children and comics.
For Ele-Fashion, there are 12 categories such as shoes, jewelry and dresses. For \textit{evaluation metrics}, we report accuracy, as it's the most common metric for node classification\cite{hu2020open, yan2023comprehensive}.
Nodes are randomly split into train/valid/test following 6/1/3 ratio.

\mypar{Knowledge Graph Completion.}
The task is to predict missing links in a knowledge graph given entities and relations. Similar to link prediction, the evaluation
is based on how well a model ranks positive test edges over negative test edges.  Specifically, we use the splits and negatives designed in ~\cite{safavi2020codex} to ensure the scope and level of difficulty of the task. Entities without multimodal information (i.e., entities that do not have descriptions or images) are filtered out. For \textit{evaluation metrics}, we report MRR, Hits@10, Hits@3 and Hits@1. 

\subsection{Statistics}

We present detailed statistics for each dataset in \cref{tab:datasets-stats} and \cref{tab:kg-stats}, and additional descriptions of each metric and property are provided in the Appendix. These statistics provide valuable insights into the structural properties of the graphs.

Our multimodal graph benchmark encompasses datasets spanning various scales and characteristics. The datasets range from small-scale (MM-CoDEx-s) to large-scale (Goodreads-NC) with varying entity and relation counts. The diversity in graph sizes, densities, and average degrees provides a comprehensive testbed for evaluating multimodal graph learning approaches across different computational complexities.

\section{Evaluation Framework}

To enable a comprehensive evaluation on multimodal graph data, \method standardizes the GNN architectures, KGEs, feature encoders, dataloaders and evaluators used.

\subsection{Graph Neural Networks}

To enable a comprehensive evaluation of GNN architectures on multimodal graph data, \method includes five representative GNN models: GCN~\cite{kipf2016semi}, SAGE~\cite{hamilton2017inductive}, MMGCN~\cite{wei2019mmgcn}, MGAT~\cite{tao2020mgat}, and BUDDY~\cite{chamberlain2022graph} . Additionally, following ~\cite{hu2020open}, we report the performance of an MLP as a baseline to evaluation the usefulness of graph structure.

\mypar{Conventional GNNs.} We standardize GCN, SAGE, MLP for both link prediction and node classification. Since BUDDY is specifically designed for link prediction, we do not report its performance on node classification. For all unimodal GNNs, text and image embeddings are first concatenated together and then passed through GNNs.

\mypar{Multimodal GNNs.}
While GNNs have shown strong performance on tasks involving graphs with unimodal features, there has been limited work on developing GNNs that can effectively handle multimodal graph data. To evaluate the effectiveness of GNNs on multimodal graphs, we adapt two recent architectures from the recommendation systems domain to our standardized multimodal graph benchmarks:

(1) MMGCN constructs separate user-item graphs for each modality. The information interchange of users and items in each modality is encoded using GNNs. Modality representations are fused in the prediction layer to predict possible recommendations~\cite{wei2019mmgcn}. For link prediction (LP), we use the most common dot product decoder to decode possible links. For node classification (NC), we stack a 3-layer MLP to transform representations to a number of dimensions equal to the number of node classes.

(2) MGAT extends the popular GAT architecture and learns modality-specific node representations which are then combined using a cross-modal attention layer~\cite{tao2020mgat}. This allows MGAT to weigh different modalities based on their importance. Similar to MMGCN, we decode links using dot product for LP and perform NC through a 3-layer MLP.

\subsection{Knowledge Graph Embeddings}

\method includes two state-of-the-art multimodal Knowledge Graph Embeddings (KGEs): MoSE~\cite{zhao2022mose} and VISTA~\cite{lee2023vista}. MoSE introduces modality-split relation embeddings for each modality, moving away from a single modality-shared embedding, which effectively mitigates modality interference. VISTA innovatively incorporates visual and textual representations of entities and relations through sophisticated transformers, utilizing entity encoding, relation encoding, and triplet decoding. Since VISTA's architecture scores entities rather than triples, we report performance using the standard 1vsAll setting instead of hard negatives~\cite{lee2023vista, ruffinelli2020you}.

Note that we do not report Graph Neural Network (GNN) performance for Knowledge Graph Completion (KGC), as prior research has demonstrated that GNN architectures can inadvertently conflate scoring functions in Knowledge Graph Embedding, potentially degrading overall performance~\cite{li2023message}.

\subsection{Feature Encoders}

\mypar{Text encoders.} For text encoders, we select CLIP~\cite{radford2021learning}, T5~\cite{raffel2020exploring}, and ImageBind~\cite{girdhar2023imagebind}. T5 is chosen as the state-of-the-art (SOTA) text embedding model, while CLIP and ImageBind are selected to ensure that the output text representations align with visual representations. Specifically, CLIP provides a multi-modal embedding approach that jointly trains text and image encoders to maximize cosine similarity between text and image embeddings. ImageBind extends this concept by enabling cross-modal representation learning across multiple modalities.

\begin{table*}[t!]
\centering
\caption{ \textbf{Link prediction results on Amazon-Sports, Amazon-Cloth and Goodreads-LP.}  Highlighted box indicates the best performing combination for each dataset. Conventional GNNs such as SAGE performs best across datasets. Aligned feature embeddings, e.g. CLIP and ImageBind outperforms unaligned features.}
\label{tab:lp_results}
\resizebox{\textwidth}{!}
{
\begin{tabular}{@{}lll c ccc c ccc c ccc@{}}
\toprule
& \multicolumn{2}{c}{Encoder} && \multicolumn{3}{c}{Amazon-Sports} & & \multicolumn{3}{c}{Amazon-Cloth}  & & \multicolumn{3}{c}{Goodreads-LP}\\
\cline{2-3} \cline{5-7} \cline{9-11} \cline{13-15}
& Image & Text && \textbf{MRR $\uparrow$} & \textbf{H@1 $\uparrow$} & \textbf{H@10 $\uparrow$} && \textbf{MRR $\uparrow$} & \textbf{H@1 $\uparrow$} & \textbf{H@10 $\uparrow$} && \textbf{MRR $\uparrow$} & \textbf{H@1 $\uparrow$} & \textbf{H@10 $\uparrow$}\\ 

\cmidrule{1-4} \cmidrule{5-8} \cmidrule{9-12} \cmidrule{13-15}

\multirow{4}{*}{MMGCN} & \multicolumn{2}{l}{CLIP} && \colorbox{lblue}{\numstd{31.96}{0.10}} & \numstd{16.35}{0.11} & \colorbox{lblue}{\numstd{68.46}{0.08}} & & \numstd{22.20}{0.05} & \numstd{10.76}{0.1} & \numstd{46.62}{0.12} & & \colorbox{lorange}{\numstd{31.84}{0.09}} & \numstd{18.63}{0.31} & \colorbox{lorange}{\numstd{59.85}{0.19}} \\
& ViT & T5 && \numstd{30.33}{0.03} & \numstd{15.01}{0.05} & \numstd{66.41}{0.11} & & \numstd{19.45}{0.34} & \numstd{9.22}{0.20} & \numstd{40.49}{0.61}& & \numstd{31.11}{0.25} & \colorbox{lorange}{\numstd{19.30}{0.45}} & \numstd{56.24}{0.19} \\
& \multicolumn{2}{l}{ImageBind} && \numstd{31.74}{0.21} & \colorbox{lblue}{\numstd{16.45}{0.13}} & \numstd{67.39}{0.74} & & \colorbox{lgreen}{\numstd{24.72}{0.19}} & \colorbox{lgreen}{\numstd{12.47}{0.09}} & \colorbox{lgreen}{\numstd{51.32}{0.56}} & & \numstd{26.32}{0.23} & \numstd{16.05}{0.22} & \numstd{46.37}{0.66}  \\
& DINOv2 & T5 && \numstd{30.04}{0.27} & \numstd{14.98}{0.07} & \numstd{64.56}{0.56} & & \numstd{21.77}{0.23} & \numstd{10.47}{0.12} & \numstd{45.81}{0.52} & & \numstd{27.64}{0.95} & \numstd{16.21}{0.65} & \numstd{51.46}{1.71} \\
\midrule
\multirow{4}{*}{MGAT} & \multicolumn{2}{l}{CLIP}  && \numstd{27.56}{0.30} & \numstd{13.55}{0.29} & \numstd{60.21}{0.21} & & \numstd{21.38}{0.23} & \numstd{10.39}{0.22} & \numstd{44.60}{0.36} & & \numstd{74.75}{1.23} & \numstd{64.53}{1.48}
& \numstd{92.81}{0.64} \\
& ViT & T5 && \colorbox{lblue}{\numstd{30.15}{0.34}} & \numstd{15.28}{0.34} & \colorbox{lblue}{\numstd{64.84}{0.41}} & & \numstd{20.59}{0.41} & \numstd{9.79}{0.30} & \numstd{43.44}{0.76} & & \colorbox{lorange}{\numstd{75.26}{1.21}} & \colorbox{lorange}{\numstd{65.23}{1.62}} & \numstd{92.90}{1.89} \\
& \multicolumn{2}{l}{ImageBind}  && \colorbox{lblue}{\numstd{30.15}{0.12}} & \colorbox{lblue}{\numstd{15.50}{0.05}} & \numstd{64.20}{0.43} & & \colorbox{lgreen}{\numstd{22.13}{0.27}} & \colorbox{lgreen}{\numstd{10.96}{0.15}} & \colorbox{lgreen}{\numstd{45.84}{0.57}} & & \numstd{74.77}{0.49} & \numstd{64.95}{0.61} & \numstd{92.51}{0.47}  \\
& DINOv2 & T5 && \numstd{28.91}{0.09} & \numstd{14.47}{0.18} & \numstd{62.11}{0.22} & & \numstd{21.42}{0.13} & \numstd{10.38}{0.13} & \numstd{44.11}{0.50} & & \numstd{74.89}{1.46} & \numstd{64.70}{1.98} & \colorbox{lorange}{\numstd{92.92}{0.41}} \\
\midrule
\multirow{4}{*}{GCN} & \multicolumn{2}{l}{CLIP} && \numstd{31.38}{0.08} & \numstd{16.58}{0.13} & \colorbox{lblue}{\numstd{66.14}{0.08}} & & \numstd{22.28}{0.05} & \numstd{11.83}{0.04} & \numstd{43.52}{0.10} & & \numstd{25.34}{0.06} & \numstd{13.81}{0.12}& \numstd{50.36}{0.14} \\
& ViT & T5 && \numstd{30.83}{0.07} & \numstd{16.31}{0.08} & \numstd{64.76}{0.15}& & \numstd{21.60}{0.05} & \numstd{11.37}{0.03} & \numstd{42.29}{0.14} & & \numstd{26.50}{0.10} & \numstd{14.86}{0.08} & \numstd{51.54}{0.14}\\
& \multicolumn{2}{l}{ImageBind}  && \colorbox{lblue}{\numstd{31.67}{0.09}} & \colorbox{lblue}{\numstd{17.07}{0.14}} & \numstd{65.61}{0.10} & & \colorbox{lgreen}{\numstd{22.81}{0.03}} & \colorbox{lgreen}{\numstd{12.27}{0.05}} & \colorbox{lgreen}{\numstd{44.28}{0.09}} & & \numstd{27.56}{1.26} & \numstd{14.31}{1.37} & \numstd{57.25}{0.52} \\
& DINOv2 & T5 && \numstd{30.42}{0.02} & \numstd{16.02}{0.03} & \numstd{64.02}{0.06} & & \numstd{21.19}{0.08} & \numstd{11.09}{0.06} & \numstd{41.46}{0.16} & & \colorbox{lorange}{\numstd{28.21}{1.12}} & \colorbox{lorange}{\numstd{15.11}{1.06}} & \colorbox{lorange}{\numstd{57.94}{0.95}}\\
\midrule
\multirow{4}{*}{SAGE} & \multicolumn{2}{l}{CLIP}  && \numstd{33.83}{0.08} & \numstd{17.57}{0.14} & \numstd{71.90}{0.07} & & \numstd{24.58}{0.18} & \numstd{12.16}{0.11} & \numstd{51.12}{0.09} & & \numstd{44.10}{1.37} & \numstd{32.32}{1.38} & \numstd{69.07}{1.19}\\
& ViT & T5 && \numstd{32.01}{0.10} & \numstd{15.94}{0.17} & \numstd{69.84}{0.21} & & \numstd{23.11}{0.05} & \numstd{11.10}{0.04} & \numstd{48.89}{0.09} & & \numstd{44.79}{0.18} & \numstd{33.11}{0.21} & \numstd{69.43}{0.18}\\
& \multicolumn{2}{l}{ImageBind}  && \colorbox{lblue}{\numstd{34.32}{0.11}} & \colorbox{lblue}{\numstd{17.87}{0.23}} & \colorbox{lblue}{\numstd{73.04}{0.15}} & & \colorbox{lgreen}{\numstd{25.20}{0.09}} & \colorbox{lgreen}{\numstd{12.63}{0.05}} & \colorbox{lgreen}{\numstd{52.53}{0.21}} & & \numstd{34.61}{0.43} & \numstd{23.82}{0.51} & \numstd{56.67}{0.21} \\
& DINOv2 & T5 && \numstd{32.20}{0.12} & \numstd{16.19}{0.2} & \numstd{69.98}{0.32}& & \numstd{22.98}{0.01} & \numstd{11.12}{0.04} & \numstd{48.28}{0.11} & & \colorbox{lorange}{\numstd{45.61}{0.22}} & \colorbox{lorange}{\numstd{34.01}{0.27}} & \colorbox{lorange}{\numstd{70.01}{0.11}}\\
\midrule
\multirow{4}{*}{BUDDY} & \multicolumn{2}{l}{CLIP}  && \numstd{31.55}{0.13}& \numstd{15.05}{0.43}& \colorbox{lblue}{\numstd{70.92}{0.25}} & & \numstd{23.44}{0.26} & \numstd{11.06}{0.20} & \numstd{51.08}{0.5} & & \colorbox{lorange}{\numstd{43.25}{0.23}} & \colorbox{lorange}{\numstd{31.84}{0.35}} & \numstd{67.93}{0.03} \\
& ViT & T5 && \numstd{30.41}{0.40} & \numstd{14.11}{0.28} & \numstd{69.55}{0.80} & & \numstd{22.82}{0.19} & \numstd{10.24}{0.12} & \numstd{51.04}{0.39} & & \numstd{43.18}{0.53} & \numstd{31.73}{0.54} & \numstd{67.89}{0.57} \\
& \multicolumn{2}{l}{ImageBind}  && \colorbox{lblue}{\numstd{33.02}{0.44}} & \colorbox{lblue}{\numstd{17.61}{0.43}} & \numstd{69.17}{0.43} & & \colorbox{lgreen}{\numstd{24.35}{0.24}} & \colorbox{lgreen}{\numstd{12.05}{0.46}} & \colorbox{lgreen}{\numstd{51.44}{0.87}} & & \numstd{41.56}{0.61} & \numstd{29.89}{0.91} & \numstd{67.41}{0.05} \\
& DINOv2 & T5 && \numstd{30.02}{0.34}& \numstd{13.78}{0.19} & \numstd{69.18}{0.67}& &\numstd{22.95}{0.06} & \numstd{10.45}{0.09} & \numstd{50.87}{0.61} & & \colorbox{lorange}{\numstd{43.25}{0.13}} & \numstd{31.77}{0.33} & \colorbox{lorange}{\numstd{68.08}{0.42}}\\
\midrule
\multirow{4}{*}{MLP} & \multicolumn{2}{l}{CLIP}  && \numstd{28.22}{0.09} & \numstd{14.54}{0.16}& \numstd{59.40}{0.08} & & \numstd{21.10}{0.04} & \numstd{10.70}{0.03} & \numstd{42.77}{0.05} & & \numstd{11.03}{0.06}& \colorbox{lorange}{\numstd{4.87}{0.04}} & \numstd{21.61}{0.11}\\
& ViT & T5 && \numstd{24.81}{0.05} & \numstd{11.63}{0.05} & \numstd{54.78}{0.04}& & \numstd{17.65}{0.06} & \numstd{8.14}{0.04} & \numstd{36.77}{0.06} & & \colorbox{lorange}{\numstd{11.10}{0.17}} & \numstd{4.84}{0.15} & \colorbox{lorange}{\numstd{21.94}{0.24}} \\
& \multicolumn{2}{l}{ImageBind}  && \colorbox{lblue}{\numstd{30.45}{0.14}} & \colorbox{lblue}{\numstd{15.91}{0.10}} & \colorbox{lblue}{\numstd{64.10}{0.07}} & & \colorbox{lgreen}{\numstd{22.18}{0.02}} & \colorbox{lgreen}{\numstd{11.42}{0.04}} & \colorbox{lgreen}{\numstd{44.86}{0.06}} & & \numstd{7.73}{0.06} & \numstd{3.37}{0.07} & \numstd{13.26}{0.03} \\
& DINOv2 & T5 && \numstd{24.81}{0.16} & \numstd{11.62}{0.18} & \numstd{54.97}{0.22} & & \numstd{17.53}{0.11} & \numstd{8.07}{0.09} & \numstd{36.53}{0.26} & &\numstd{10.28}{0.04} & \numstd{4.49}{0.05} & \numstd{19.86}{0.03} \\

\bottomrule
\end{tabular}
}
\end{table*}

\mypar{Visual encoders.} For visual encoders, we select CLIP~\cite{radford2021learning}, ViT~\cite{dosovitskiy2020image}, ImageBind~\cite{girdhar2023imagebind}, and DINOv2~\cite{oquab2023dinov2}. ViT and DINOv2 represent two distinct approaches to visual feature learning: ViT is explicitly trained with supervision, while DINOv2 learns robust visual features through self-supervised methods. CLIP and ImageBind are chosen to explore text-visual representation alignment. Notably, ImageBind demonstrates potential for extending multimodal graph learning across diverse modalities like audio and video by embedding them in a unified space.

Our selection of feature encoders enables us to investigate critical multimodal graph learning design, specifically:

\begin{itemize}
    \item The importance of aligning multiple modalities into a unified embedding space.
    \item Comparative performance of supervised versus unsupervised feature encoders.
\end{itemize}

\subsection{Dataloader and Evaluators}

\method also provides a standardized dataloader and evaluator implementation. The dataloader is built upon PyTorch's DataLoader class. The evaluator is based on PyTorch's evaluator and performs standardized and reliable evaluation. The dataloader supports various sampling strategies, such as neighbor sampling  and layer-wise sampling, to scale the training process to large graphs. It also enables distributed training using PyTorch's distributed data parallel (DDP)  for faster training on multiple GPUs.

By standardizing the GNNs, KGEs, feature encoders, dataloader and evaluators, \method ensures a fair and comprehensive evaluation of graph learning algorithms on multimodal graph data.

\section{Empirical Analysis}
\label{sec:experiments}

We conduct a comprehensive evaluation of our proposed benchmarks on all tasks. Our experiments aim to answer the following research questions:

\noindent \textbf{(RQ1)} What are the performances of MM-GNNs, GNNs, and MM-KGEs on \method?

\noindent \textbf{(RQ2)} Do multimodal-GNNs consistently outperform conventional GNNs on \method?

\noindent \textbf{(RQ3)} What is the most effective approach for encoding features, and is alignment between modalities necessary?

\noindent \textbf{(RQ4)} How much performance gain can be achieved by using multimodal features compared to unimodal features alone?

\begin{figure}[t!]
\centering
  \includegraphics[width=0.8\linewidth]{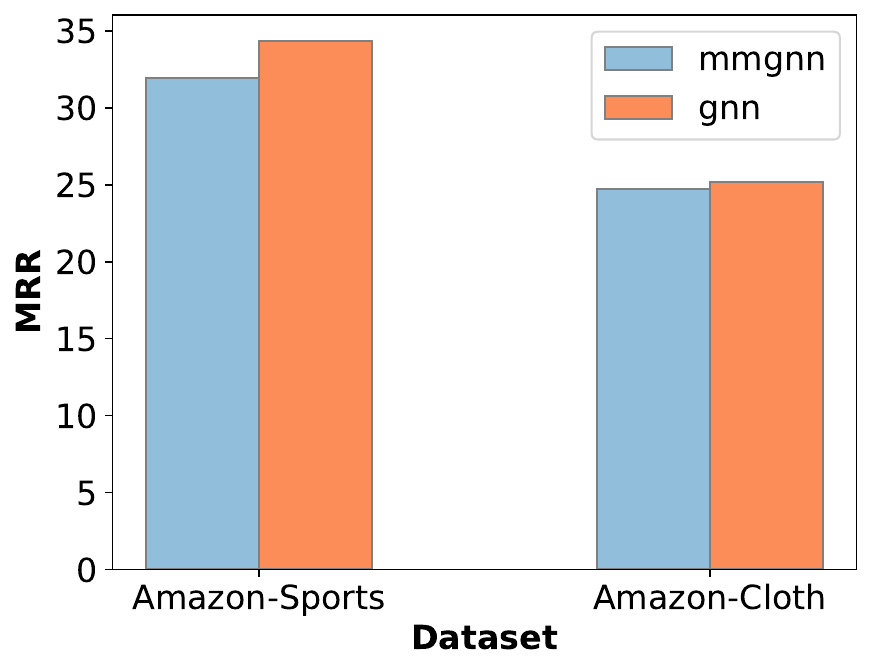}
  \vspace{-0.4cm}
  \caption{\textbf{(RQ2) Multimodal GNNs underperforms conventional GNNs.} We compare the best performance of multimodal GNNs (MMGCN/MGAT) and conventional GNNs (SAGE, GCN, BUDDY). Conventional GNNs consistently perform better across datasets, which justifies the importance of building \method~ and calls for better multimodal GNN designs. }
  \label{fig:mmgnn}
  \vspace{-0.3cm}
\end{figure}

To ensure fair and rigorous comparisons between various feature encoders and GNNs, we adopt consistent experimental settings across all experiments. We perform automatic hyperparameter tuning using Optuna~\cite{optuna_2019}  to optimize the performance of each model. Detailed experimental setup, hyperparameters and additional experiment results are provided in the Appendix.

\subsection{Results}

\textbf{(RQ1)} The detailed link prediction results are shown in \cref{tab:lp_results}. 
The detailed node classification results are shown in \cref{tab:nc_results}. The detailed knowledge graph completion results  are shown in \cref{tab:kg_mose}. 
\begin{table*}[t!]
\centering

\captionof{table}{\textbf{KGC results on MM-Codex-S and MM-Codex-M.} Highlighted box indicates the best performing combination for each dataset. Aligned feature embeddings outperforms unaligned features.} 
\label{tab:kg_mose}
\resizebox{0.7\textwidth}{!}
{
\begin{tabular}{@{}llll ccc c@{}}
\toprule
KGEs & Dataset & Image Encoder & Text Encoder & MRR & H@1 &  H@3 & H@10\\
\cmidrule{1-2} \cmidrule{3-4} \cmidrule{5-8}
\multirow{8}{*}{MoSE} & \multirow{4}{*}{MM-CoDEx-s} & CLIP & CLIP & \numstd{36.59}{0.16} & \numstd{28.10}{0.51}  & \numstd{41.79}{1.03}  & \numstd{51.08}{0.29} \\
& & T5 & ViT & \numstd{39.00}{0.35}  & \numstd{34.40}{0.47} & \numstd{42.73}{0.33}  & \numstd{46.53}{0.49} \\
& & ImageBind & ImageBind& \numstd{22.45}{1.36}  & \numstd{5.78}{1.33} & \numstd{35.48}{1.30}  & \numstd{44.49}{0.98} \\
& & T5 & DINOv2 & \colorbox{Gray}{\numstd{40.07}{0.24}}  & \colorbox{Gray}{\numstd{35.82}{1.85}} & \colorbox{Gray}{\numstd{42.77}{1.17}}  & \colorbox{Gray}{\numstd{47.30}{0.17}} \\
\cmidrule{2-8} 
& \multirow{4}{*}{MM-CoDEx-m} & CLIP & CLIP & \numstd{6.97}{0.58}  & \colorbox{Gray}{\numstd{3.24}{1.11}} & \numstd{9.33}{0.33} & \numstd{12.15}{0.02}  \\
& & T5 & ViT & \numstd{6.09}{0.01}  & \numstd{1.71}{0.22} & \numstd{8.42}{0.26}  & \numstd{12.56}{0.28} \\
& & ImageBind & ImageBind& \colorbox{Gray}{\numstd{7.01}{0.15}}  & \numstd{2.26}{0.33} & \colorbox{Gray}{\numstd{9.60}{0.26}} & \colorbox{Gray}{\numstd{13.98}{0.24}} \\
& & T5 & DINOv2 & \numstd{6.40}{0.18}  & \numstd{2.14}{0.22} & \numstd{9.06}{0.23}  & \numstd{12.74}{0.14} \\
\midrule
\multirow{8}{*}{VISTA} & \multirow{4}{*}{MM-CoDEx-s} & CLIP & CLIP & \numstd{28.49}{0.39} & \numstd{19.12}{0.64}  & \numstd{30.53}{0.46}  & \numstd{49.83}{0.31} \\
& & T5 & ViT & \numstd{29.70}{0.51}  & \numstd{20.00}{0.59} & \numstd{32.11}{0.31}  & \numstd{50.52}{0.75} \\
& & ImageBind & ImageBind& \colorbox{Gray}{\numstd{30.39}{0.65}}  & \colorbox{Gray}{\numstd{20.20}{0.84}} & \colorbox{Gray}{\numstd{33.52}{0.30}}  & \colorbox{Gray}{\numstd{52.68}{0.50}} \\
& & T5 & DINOv2 & \numstd{26.98}{1.68}  & \numstd{17.81}{1.16} & \numstd{29.53}{2.58}  & \numstd{47.28}{2.86} \\
\cmidrule{2-8} 
& \multirow{4}{*}{MM-CoDEx-m} & CLIP & CLIP & \numstd{22.19}{1.92}  & \numstd{15.81}{1.62} & \numstd{24.31}{2.10} & \numstd{34.73}{2.58}  \\
& & T5 & ViT & \numstd{22.10}{1.61}  & \numstd{15.92}{1.33} & \numstd{24.09}{1.71}  & \numstd{34.68}{2.26} \\
& & ImageBind & ImageBind& \colorbox{Gray}{\numstd{23.20}{1.45}}  & \colorbox{Gray}{\numstd{16.97}{1.35}} & \colorbox{Gray}{\numstd{24.95}{1.24}} & \colorbox{Gray}{\numstd{35.88}{1.91}} \\
& & T5 & DINOv2 & \numstd{21.38}{1.17}  & \numstd{15.35}{1.15} & \numstd{23.19}{1.06}  & \numstd{33.40}{1.49} \\
\bottomrule
\end{tabular}
}

\end{table*}
Here are our findings:

\mypar{(RQ2) Multimodal GNNs perform poorly, while GNNs and MM-KGES often outperforms on the proposed \method datasets.} Our experiments, as illustrated in \cref{fig:mmgnn} and \cref{tab:lp_results}, \cref{tab:nc_results}, and \cref{tab:kg_mose}, reveal an unexpected trend: multimodal GNNs specifically designed for processing multimodal input data, such as MMGCN and MGAT, do not consistently outperform conventional GNN models like SAGE across various tasks and datasets.
This counterintuitive result may be attributed to the architectural designs of MMGCN and MGAT, which primarily perform message passing and aggregation on each modality separately, only fusing information across modalities at the final stage. We hypothesize that this approach leads to insufficient integration or alignment of multimodal information, highlighting the need for more sophisticated model designs for multimodal graphs.

\begin{figure}[t!]
	\centering
    \subfloat{%
      \includegraphics[width=.50\textwidth]{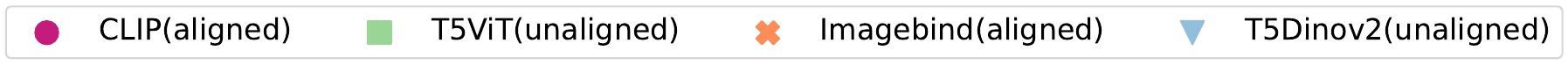} 
    }
	\setcounter{subfigure}{0}
	
    \subfloat[Amazon-Sports\label{AS}]{%
      \includegraphics[width=0.3\textwidth]{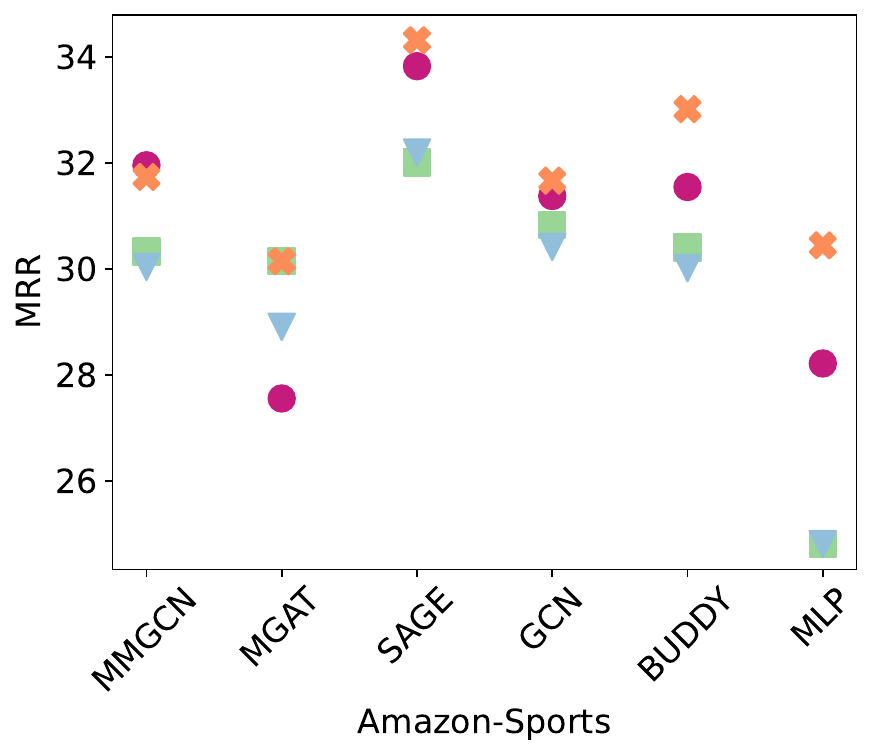}
    }\\
    ~
    \subfloat[Amazon-Cloth\label{AC}]{%
      \includegraphics[width=0.3\textwidth  
 ]{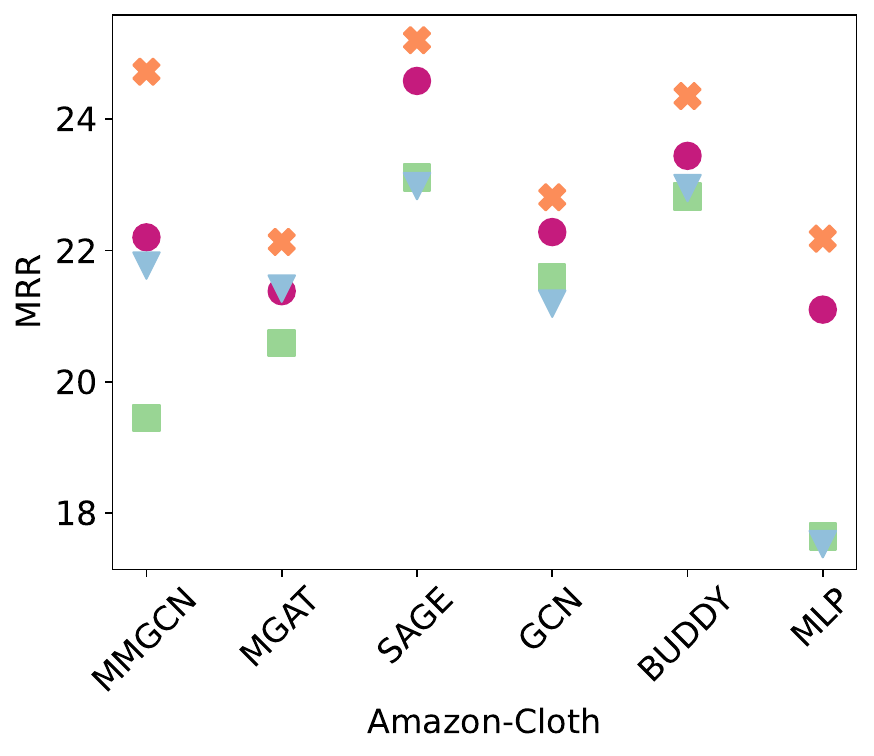}
    }
    ~
	\centering
\caption{\textbf{(RQ3) Feature alignment is important.} We compare the performance of various feature encoders, find aligned features, e.g., CLIP and Imagebind result in much better performance compared with unaligned features on Amazon-Sports and Amazon-Cloth. Among them, Imagebind performs the best across backbones. This indicates the importance of using aligned features on these datasets.}
    \label{fig:align}
    \vspace{-1em}
\end{figure}

In contrast, we observed that MoSE and VISTA perform notably well on the two proposed MMKG datasets, with the exception of MoSE on MM-CoDEx-m. MoSE's approach of learning separate KGE embeddings for each modality and then ensembling them together represents a form of late fusion. While this method shows promise, its performance variability across datasets suggests that the effectiveness of late fusion strategies may be dataset-dependent.

Our new findings underscore the complexity of effectively leveraging multimodal information in graph learning tasks. They also highlight the need for further research into more advanced fusion techniques that can better capture and integrate the complementary information provided by different modalities throughout the learning process, rather than relying on late-stage integration.

\mypar{(RQ3) Feature alignment is important for multimodal graph learning.}
Our evaluation of four feature encoders, CLIP, ImageBind, T5 + ViT, and T5 + DINOv2, reveals significant performance variations across datasets and tasks. CLIP and ImageBind, which map features from various modalities to a shared embedding space, consistently outperform T5 + ViT and T5 + DINOv2, which employ independent embedding methods for each modality without specific alignment layers.

\begin{table}[t!]
\centering
\caption{\textbf{Node classification results on Goodreads-NC and Ele-Fashion.} Conventional GNNs such as SAGE performs best across datasets. Aligned feature embeddings, e.g. CLIP and
ImageBind outperforms unaligned features. } 
\label{tab:nc_results}
\resizebox{0.5 \textwidth}{!}
{
\begin{tabular}{@{}lll c c@{}}
\toprule
& Image Encoder & Text Encoder & Ele-fashion & Goodreads-NC\\
\cmidrule{1-2} \cmidrule{3-4} \cmidrule{5-5}
\multirow{4}{*}{MMGCN} & CLIP & CLIP & \numstd{86.10}{0.50} & \textbf{\numstd{83.29}{0.20}} \\
& T5 & ViT & \numstd{82.39}{0.30} & \numstd{81.85}{0.22}\\
& ImageBind & ImageBind & \numstd{86.21}{0.94} & \numstd{80.58}{1.08}\\
& T5 & DINOv2 & \numstd{85.53}{0.33} & \numstd{82.44}{0.11}\\
\midrule
\multirow{4}{*}{MGAT} & CLIP & CLIP & \numstd{84.66}{0.29} & \numstd{76.48}{0.59}\\
& T5 & ViT & \numstd{84.01}{0.08} & \numstd{75.43}{0.76}\\
& ImageBind & ImageBind & \numstd{86.12}{0.08} & \numstd{69.45}{6.25}\\
& T5 & DINOv2 & \numstd{84.54}{0.27} & \numstd{74.98}{1.23}\\
\midrule
\multirow{4}{*}{GCN} & CLIP & CLIP & \numstd{79.83}{0.03} & \numstd{81.61}{0.01}\\
& T5 & ViT & \numstd{79.63}{0.07} & \numstd{81.67}{0.03}\\
& ImageBind & ImageBind & \numstd{80.35}{0.02} & \numstd{78.91}{0.04}\\
& T5 & DINOv2 & \numstd{79.37}{0.04} & \numstd{81.71}{0.03}\\
\midrule
\multirow{4}{*}{SAGE} & CLIP & CLIP & \numstd{87.10}{0.02} & \colorbox{Gray}{\numstd{83.30}{0.02}}\\
& T5 & ViT & \numstd{84.41}{0.09} & \numstd{83.03}{0.04}\\
& ImageBind & ImageBind & \textbf{\numstd{87.71}{0.13}} & \numstd{80.39}{0.21}\\
& T5 & DINOv2 & \numstd{85.31}{0.09} & \numstd{82.99}{0.08}\\
\midrule
\multirow{4}{*}{MLP} & CLIP & CLIP& \numstd{85.16}{0.03} & \numstd{72.29}{0.02}\\
& T5 & ViT & \numstd{84.98}{0.05} & \numstd{67.82}{0.07}\\
& ImageBind & ImageBind & \colorbox{Gray}{\numstd{88.73}{0.01}} & \numstd{58.75}{0.05}\\
& T5 & DINOv2 & \numstd{84.87}{0.01} & \numstd{68.83}{0.03}\\
\bottomrule
\end{tabular}
}
\vspace{-0.3cm}
\end{table}

\begin{figure}[t]
	\setcounter{subfigure}{0}
    ~
    \subfloat[Link Prediction\label{lp}]{%
      \includegraphics[width=0.35\textwidth]{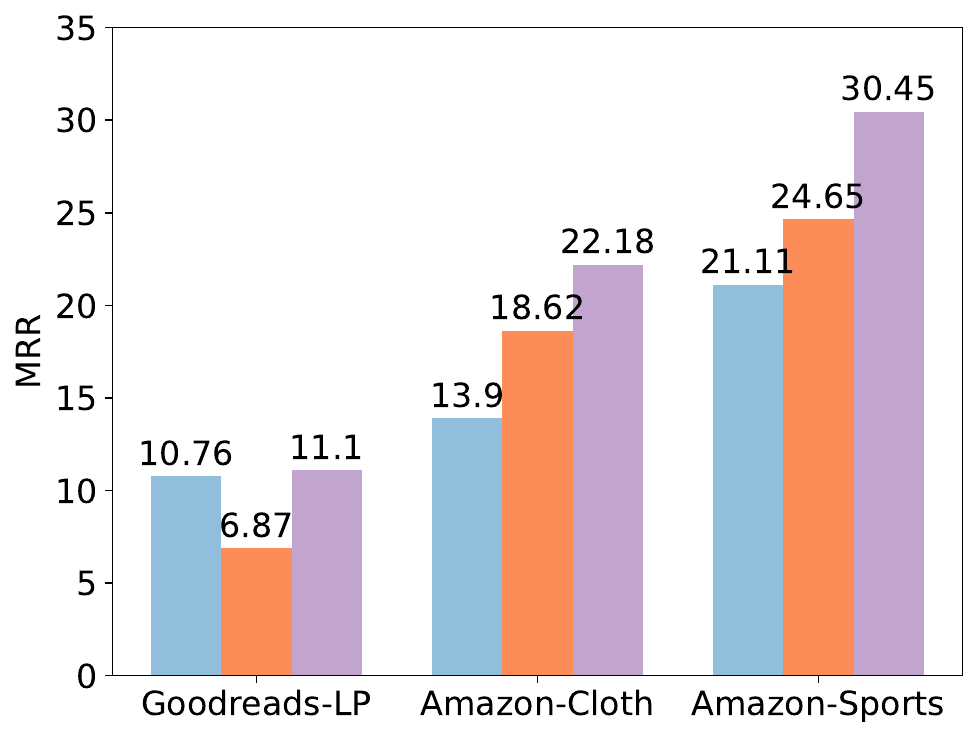}
    } \\
    ~
    \subfloat[Node Classification
    \label{nc}]{\includegraphics[width=0.35\textwidth]{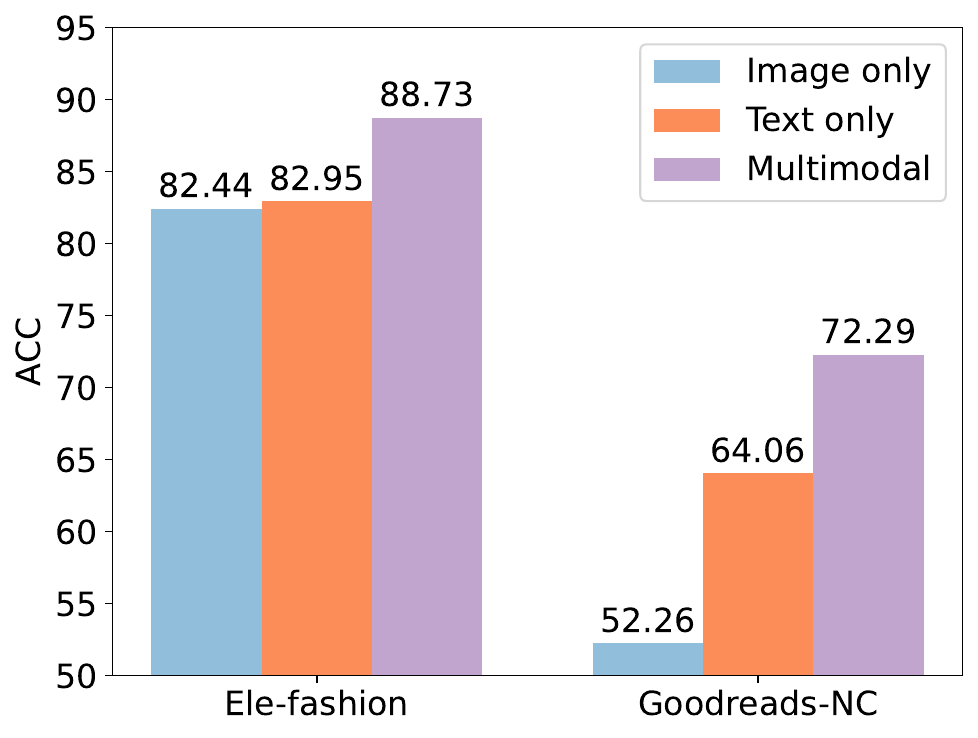}
    }
	\centering
\caption{\textbf{(RQ4) Multimodal features are helpful for graph learning.} Multimodal features perform better than text-only features across datasets and tasks, which justifies the necessity of introducing multimodal graph datasets.}
    \label{fig:text-only}
    \vspace{-1em}
\end{figure}

For \textit{link prediction}, aligned multimodal features consistently outperform unaligned features, as illustrated in \cref{fig:align}. ImageBind yields the best results across all backbones. However, on the Goodreads-LP dataset, T5+DINOv2 demonstrates superior performance due to its robust Optical Character Recognition (OCR) capabilities, which are crucial for extracting textual information from book covers. In \textit{node classification} tasks, ImageBind and CLIP again emerge as the top-performing feature encoders. This further emphasizes the importance of feature alignment in graph-based tasks. Similarly, for \textit{knowledge graph completion}, ImageBind consistently demonstrates superior performance across datasets and various KGEs, except for MM-CoDEx-s on MoSE. For MM-CoDEx-s on MoSE, we suspect that the modality split function in MoSE leads to unaligned embedding space for KGEs and thus hurts the performance.

On Goodreads-LP, the T5 + DINOv2 combination demonstrates superior performance.
We believe it is because the Goodreads dataset presents a distinctive challenge because book covers primarily contain textual information, requiring robust capabilities of extracting and understanding text from book covers. This is crucial because author information is important when predicting reading preferences, and it is only displayed on book covers. Users tend to prefer books by the same author, making OCR-capable models particularly effective in this context. Our findings align with Tong et al.~\cite{tong2024eyes}, where CLIP-based VLM is shown to be suboptimal at capturing details.

\mypar{(RQ4) Multimodal models outperform unimodal models, underscoring the importance of visual information.} To highlight the necessity of multimodal graph benchmarks beyond existing text-rich graphs, we conducted a study comparing the performance gains achieved by multimodal features over unimodal text or visual features alone (T5/ViT). Using an MLP model with optimal hyperparameters selected through Optuna, we focused on isolating the effects of feature types. As shown in \cref{fig:text-only}, integrating multimodal features consistently results in more than a 6\% improvement in performance across various datasets and tasks. While visual information alone does not surpass text in performance, the combination of both modalities significantly outperforms any single modality. This finding underscores the critical role of visual information in enhancing graph representations. Note that we do not report results for knowledge graph completion (KGC) tasks, as all evaluated KGEs require multimodal inputs.

\section{Conclusion}
\label{sec:conclusion}

This paper presents \method, a comprehensive benchmark for multimodal graph learning. \method includes seven diverse datasets with rich textual and visual features, standardizing GNN architectures, KGEs, feature encoders, dataloaders, and evaluators to ensure fair evaluation. This setup provides insights into the performance of various GNN and encoder combinations in multimodal settings.
While \method advances the integration of text and visual data, challenges remain in incorporating audio and video features due to the scarcity of open-source datasets. Although ImageBind features offer a basis for future research, further exploration is needed to effectively integrate these modalities.
\method aims to drive the development of robust graph learning algorithms for real-world applications by highlighting the importance of visual information and setting the stage for future innovations in multimodal graph learning.

\section*{Acknowledgments}
This material is based upon work supported by a SnapChat gift, the National Science Foundation under IIS~2212143 and  CAREER Grant No.~IIS 1845491. Any opinions, findings, and conclusions or recommendations expressed in this material are those of the author(s) and do not necessarily reflect the views of the National Science Foundation or other funding parties.

{
    \small
    \bibliographystyle{ieeenat_fullname}
    \bibliography{main}
}
\appendix

\section{Graph Statistics Metrics}
To gain insights into the structural properties of the graphs in each dataset, we compute the following graph statistics metrics:

\begin{itemize}
    \item \textbf{Average Degree.} This metric captures the density of the graphs, with higher values indicating denser graphs. It is calculated as the average number of edges connected to each node in the graph.
    \item \textbf{Average Clustering Coefficient (CC).} The clustering coefficient measures the tendency of nodes to form tightly connected groups or clusters. A higher clustering coefficient suggests a greater likelihood of nodes forming densely connected communities. It is computed as the average of the local clustering coefficients of all nodes in the graph.
    \item \textbf{Resource Allocation.} Resource Allocation is a commonly used heuristic for link prediction tasks. It measures the likelihood of two nodes being connected based on their shared neighbors. The performance of this heuristic on each dataset serves as an indication of how well simple heuristic-based methods may perform, providing a baseline for more sophisticated graph learning algorithms.
    \item \textbf{Transitivity.} Transitivity quantifies the probability that two nodes with a common neighbor are also connected to each other, forming a triangle. It is calculated as the ratio of the number of closed triplets (triangles) to the total number of triplets (both open and closed) in the graph.
    \item \textbf{Edge Homophily.} Edge Homophily captures the similarity of node labels with respect to the labels of their neighboring nodes. A high Edge Homophily suggests that nodes with similar labels tend to connect to each other, which can be exploited by graph learning algorithms. Edge Homophily is not applicable (N/A) to link prediction datasets, as nodes in these datasets do not have labels.
\end{itemize}

\section{Link Prediction Evaluation}

\mypar{Negative Sampling.}
For the task of link prediction, we need to evaluate the performance of models in ranking positive edges (existing edges in the graph) higher than negative edges (non-existent edges). To generate negative edges for evaluation, we adopt the widely-used approach of dedicatedly sampling a fixed number of negative edges (e.g., 150) for each positive edge. This approach has been shown to be more effective than using a single, large set of negative edges for all positive edges~\cite{li2024evaluating}.

Given a set of positive edges and their corresponding negative edges, we employ the following evaluation metrics to assess the link prediction performance:

\mypar{Hits@K.} 
This metric measures whether the true positive edge is ranked within the top K predictions made by the link prediction model. Specifically, the model scores and ranks all positive and negative edges. Hits@K is then calculated as the fraction of positive edges that are ranked among the top K predictions.

A higher Hits@K value indicates better performance in ranking positive edges among the top predictions. However, it does not provide information about the specific ranks of positive edges beyond the top K. Therefore, we also report the Mean Reciprocal Rank (MRR) metric.

\mypar{Mean Reciprocal Rank (MRR).} MRR is a widely-used metric that considers the specific ranks of positive edges in the predicted ranking. For each positive edge, the reciprocal of its rank in the predicted ranking is calculated. MRR is then computed as the mean of these reciprocal ranks over all positive edges.

A higher MRR value indicates better ranking performance, with a perfect ranking yielding an MRR of 1. MRR provides a more nuanced evaluation of the ranking quality compared to Hits@K, as it considers the entire ranking rather than just the top K predictions.

By reporting both Hits@K and MRR, we provide a comprehensive evaluation of link prediction performance on the multimodal graph datasets in \method. These metrics enable researchers to assess the strengths and weaknesses of different graph learning algorithms in ranking positive edges higher than negative edges, a crucial task in many real-world applications involving graph data.

\section{Other Candidate Datasets}
We also provide a list of candidate datasets and the reasons why they are unsuitable for the our benchmark here.

\textbf{Ogbn-Arxiv:} Raw paper IDs are not provided and cannot be retrieved by backtracking, making it impossible to crawl images from Arxiv.

\textbf{Yelp:} Only reviews of restaurants have textual and visual information. However, the graph structure of user-friend-with-user exists, and typically, a user's friends do not have any reviews on any restaurant. Additionally, Yelp's license prevents repurposing the datasets.

\textbf{Tiktok:} This dataset does not provide any raw text or image features.

\section{Experimental details}
~\label{sec:details}

To ensure a fair and comprehensive evaluation of different graph learning algorithms on the Multimodal Graph Benchmark (\method), we conduct extensive experiments with rigorous experimental settings and hyperparameter tuning. In this section, we provide detailed information about the experimental setup, hyperparameter search process, and optimization strategies employed in our study.

\subsection{Hyperparameter Tuning}

Proper hyperparameter tuning is crucial for obtaining reliable and meaningful results when evaluating machine learning models. To this end, we perform automatic hyperparameter search using Optuna~\cite{optuna_2019}, a state-of-the-art hyperparameter optimization framework. The hyperparameter search is directly optimized towards maximizing the evaluation metric of interest for each task.

Specifically, for the link prediction task, we optimize the hyperparameters to maximize the Mean Reciprocal Rank (MRR) metric, as it provides a more nuanced evaluation of ranking performance compared to Hits@K. For the node classification task, we optimize the hyperparameters to maximize the Accuracy metric.
The hyperparameter search space includes the following key hyperparameters:

\begin{itemize}
    \item \textbf{Learning Rate (LR)}: We explore a wide range of learning rates, from 1e-1 to 1e-5, to find the optimal value for each model and dataset combination.
    \item \textbf{Number of GNN Layers (\# Layers)}: We vary the number of Graph Neural Network (GNN) layers from 1 to 3, as the depth of the GNN architecture can significantly impact its performance.
\end{itemize}

To ensure robust and reliable results, we perform 20 independent hyperparameter studies for each combination of feature encoders and GNN models on a single A40 GPU. Each experiment is run three times with different random seeds, and we report the mean and standard deviation of the respective performance metrics.

\subsection{Optimization}

All experiments are run with 3 different random seeds to account for the stochasticity in model training and initialization. We report the mean and standard deviation of the respective performance metrics (MRR for link prediction and Accuracy for node classification) across the 3 runs.

For optimization, we use the Adam optimizer with default settings, which has been shown to work well for a wide range of deep learning tasks. To further improve the learning process, we employ a learning rate scheduler with a decay factor of $\gamma=0.1$ and a step size of 5 epochs. This allows the model to make larger updates in the early stages of training and fine-tune the parameters in the later stages.

\subsection{Additional Results}

\vspace{0.3em}
\noindent
{\bf SOTA methods applied to text-only data.} 
We compare ImageBind+SAGE with UniGraph, one of the SOTA methods on text-attributed graphs, and results shown in \cref{tab:unigraph} are from ~\cite{he2025unigraph2}. 
It demonstrates that multimodal inputs consistently outperform text-only inputs. 
\begin{table}[t!]
\centering
\caption{\textbf{Node classification results on Goodreads-NC and Ele-Fashion.} } 
\label{tab:unigraph}
\resizebox{0.5 \textwidth}{!}
{
\begin{tabular}{lcc}
\toprule
Dataset  & UniGraph & ImageBind+SAGE \\
\midrule
Amazon-Sports & 27.11 $\pm$ 0.10 & 34.32 $\pm$ 0.11 \\

Amazon-Cloth & 18.01 $\pm$ 0.03 & 25.20 $\pm$ 0.09 \\

Goodreads-LP  & 22.31 $\pm$ 0.05 & 34.61 $\pm$ 0.43 \\

Goodreads-NC & 78.14 $\pm$ 0.11 & 80.39 $\pm$ 0.21 \\

Ele-fashion & 81.05 $\pm$ 0.08 & 87.71 $\pm$ 0.13 \\

\bottomrule
\end{tabular}
}
\vspace{-0.3cm}
\end{table}

\paragraph{F1 score}
Additionally, we report F1-score on the two node classification datasets. The results are shown below. 

\begin{table}[t!]
\centering
\caption{\textbf{Additional results on Macro-F1 score} } 

\resizebox{0.5 \textwidth}{!}
{
\begin{tabular}{@{}lll c c@{}}
\toprule
& Image Encoder & Text Encoder & Ele-fashion & Goodreads-NC\\
\cmidrule{1-2} \cmidrule{3-4} \cmidrule{5-5}
\multirow{4}{*}{MMGCN} & CLIP & CLIP &  \numstd{75.07}{0.45} &  \numstd{81.02}{0.32}\\
& T5 & ViT & \numstd{63.98}{1.18} & \numstd{79.45}{0.49}\\
& ImageBind & ImageBind & \numstd{76.18}{0.72} & \numstd{78.41}{0.45} \\
& T5 & DINOv2 & \numstd{73.62}{0.04} & \numstd{80.13}{0.13}\\
\midrule
\multirow{4}{*}{GCN} & CLIP & CLIP & \numstd{79.85}{0.04} & \numstd{74.36}{0.01}\\
& T5 & ViT & \numstd{79.69}{0.08}  & \numstd{76.27}{0.03}\\
& ImageBind & ImageBind &\numstd{80.47}{0.12} & \numstd{72.84}{0.10}\\
& T5 & DINOv2 & \numstd{79.39}{0.09} & \numstd{81.72}{0.08}  \\
\midrule
\multirow{4}{*}{SAGE} & CLIP & CLIP & \numstd{87.08}{0.01} & \numstd{83.30}{0.04} \\
& T5 & ViT & \numstd{85.81}{0.04} & \numstd{82.94}{0.08}\\
& ImageBind & ImageBind & \numstd{87.72}{0.08} & \numstd{80.26}{0.04}\\
& T5 & DINOv2 & \numstd{85.45}{0.12} & \numstd{82.88}{0.02}\\

\bottomrule
\end{tabular}
}

\end{table}

\end{document}